\documentclass[letterpaper, 10 pt, conference]{ieeeconf}  

\IEEEoverridecommandlockouts                              

\usepackage{scalefnt}
\newcommand\msf{0.99}
\AtBeginDocument{\scalefont{\msf}}

\usepackage{cite}

\usepackage{comment}
\usepackage[T1]{fontenc}
\usepackage{diagbox} 
\usepackage{graphicx}
\usepackage{subcaption}
\usepackage{amsmath,amssymb,amsfonts}
\usepackage{xcolor}

\usepackage{float}

\makeatletter
\let\NAT@parse\undefined
\makeatother
\usepackage{hyperref}

\title{\LARGE \bf
Generating Diverse Challenging Terrains for Legged Robots Using Quality-Diversity Algorithm
}

\author{Arthur Esquerre-Pourtère$^{1}$, Minsoo Kim$^{1}$, and Jaeheung Park$^{1, 2}$
\thanks{*This work was supported by the National Research Foundation of Korea (NRF) grant, funded by the Korea government (MSIT) (No. 2021R1A2C3005914), and by the Basic Science Research Program through the NRF, funded by the Ministry of Education (RS-2023-00274280).}
\thanks{$^{1}$A. Esquerre-Pourtère, M. Kim, and J. Park are with the Department of Intelligence and Information, Graduate School of Convergence Science and Technology, Seoul National University, Seoul, Republic of Korea (\{\href{mailto:tutur263@snu.ac.kr}{tutur263}, \href{mailto:msk930512@snu.ac.kr}{msk930512}, \href{mailto:park73@snu.ac.kr}{park73}\}@snu.ac.kr).}
\thanks{$^{2}$J. Park is also with ASRI, AIIS, Seoul National University, Seoul, Republic of Korea, and the Advanced Institute of Convergence Technology, Suwon, Republic of Korea. He is the corresponding author of this paper.}
}

\begin{document}

\maketitle
\thispagestyle{empty}
\pagestyle{empty}

\begin{abstract}

While legged robots have achieved significant advancements in recent years, ensuring the robustness of their controllers on unstructured terrains remains challenging. It requires generating diverse and challenging unstructured terrains to test the robot and discover its vulnerabilities. This topic remains underexplored in the literature. This paper presents a Quality-Diversity framework to generate diverse and challenging terrains that uncover weaknesses in legged robot controllers. Our method, applied to both simulated bipedal and quadruped robots, produces an archive of terrains optimized to challenge the controller in different ways. Quantitative and qualitative analyses show that the generated archive effectively contains terrains that the robots struggled to traverse, presenting different failure modes. Interesting results were observed, including failure cases that were not necessarily expected. Experiments show that the generated terrains can also be used to improve RL-based controllers. 

\end{abstract}

\section{INTRODUCTION} 

Recent progress in legged robotics \cite{li2024reinforcement, hwangbo2019learning, kumar2021rma, miki2022learning}, particularly through the use of reinforcement learning (RL), has led to significant improvements in their performance in navigating complex terrains. However, despite these advances, significant challenges remain in ensuring the robustness of such systems, particularly when navigating unstructured terrains. 

Traversing unstructured terrains is crucial in applications that require the exploration of hazardous areas, such as rescue operations or underground inspections.
Many studies rely on hand-crafted terrains, such as stairs, slopes, and discrete obstacles, or employ uncontrollable noise or Perlin noise \cite{lagae2010survey,miki2022learning} to generate them. While these methods allow for training and testing controllers on a variety of terrains, their scope is limited, and they do not ensure the controller's reliability across all possible terrains. Critical corner cases may be missed, and, given the diversity of terrains the robot might encounter, these cases can be difficult to identify, especially as weaknesses can differ widely depending on the controller's design.

Moreover, such weaknesses are often difficult to discover manually. In \cite{shi2024rethinking}, 100 volunteers were asked to identify weaknesses in a quadruped robot's controller by applying pushing forces and overwriting velocity commands. The authors found that learned adversarial attacks were more efficient than humans in finding these weaknesses. Although weaknesses have been identified through external disturbances like pushing forces, the topic of identifying various failure cases in legged robot controllers remains insufficiently addressed in the literature, particularly in the exploration of failures across different unstructured terrains.

This paper proposes a framework based on Quality-Diversity (QD) algorithms \cite{pugh2016quality, cully2017quality} that generate challenging terrains to discover a diverse set of failure modes of a legged robot controller. The framework generates a collection (archive) of terrains that are diverse in the sense that the target controller encounters various types of failures when traversing them. Each type of failure results in a specific penalty, which is used as a descriptor to capture the diversity of the terrains, as illustrated in Fig.\ref{fig:archive_example}. The terrains are optimized to maximize the sum of the penalties, thus generating highly challenging terrains. The proposed terrain generation framework makes no assumptions about the controller and treats it as a black box. 

The main contributions are summarized as follows:


\begin{figure}[t]
    \centering
    \includegraphics[width=0.49\textwidth]{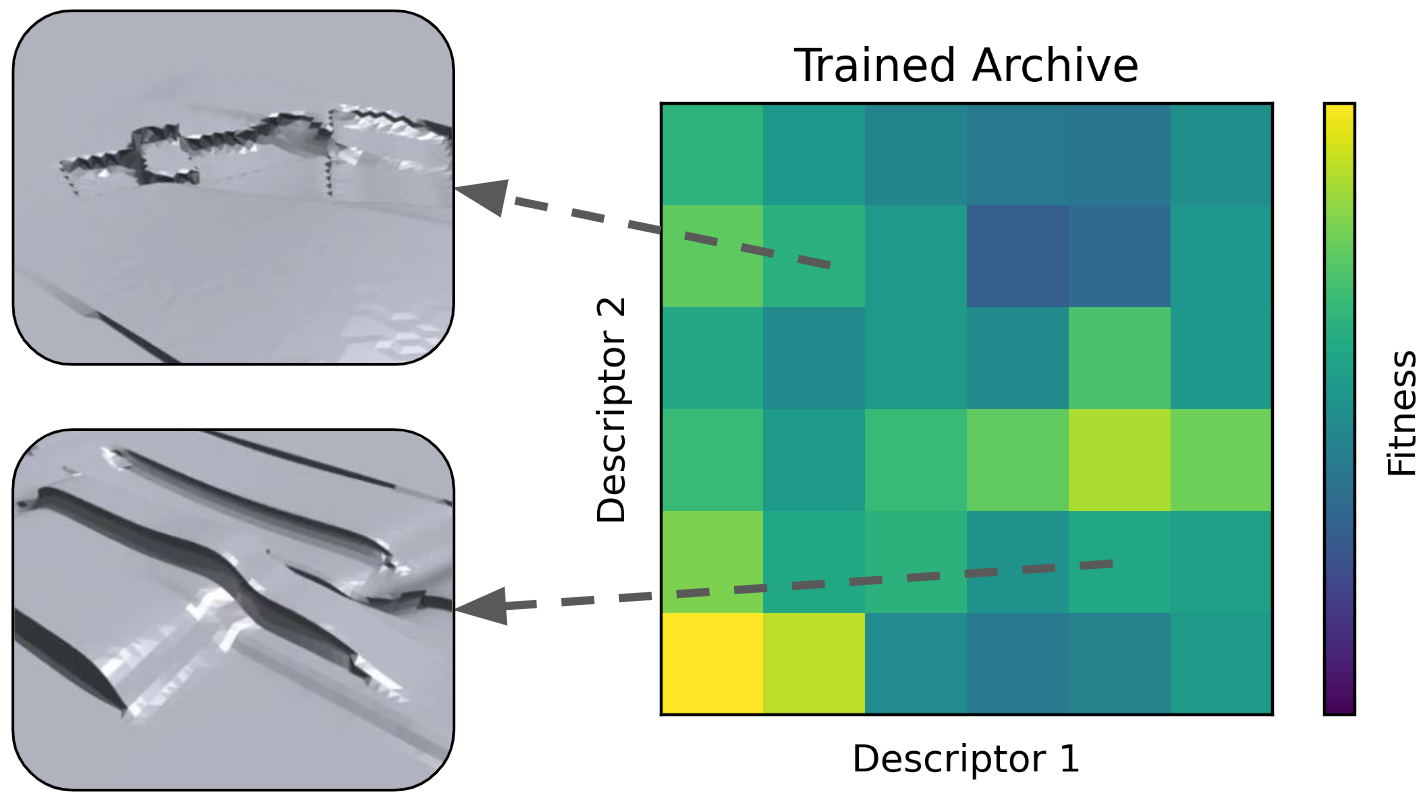}
    \caption{QD archive of terrains.}
    \label{fig:archive_example}
\end{figure}

\begin{itemize}
    \item The generation of diverse, challenging terrains is formulated as a Quality-Diversity problem. To the best of our knowledge, this is the first attempt to apply QD algorithms specifically to identify corner cases in legged robot controllers. 
    \item Experiments in simulations of both bipedal and quadruped robots demonstrate the method’s effectiveness and its ability to generate scenarios that improve the performance of RL-based controllers.

\end{itemize}

The remainder of this paper is structured as follows: Section \ref{related_works} reviews the related work. Section \ref{methodology} outlines our methodology for generating terrains via Quality-Diversity algorithms. Section \ref{exp} presents our experiments and results. Section \ref{conclusion} concludes the paper.

\section{RELATED WORKS}
\label{related_works}

\subsection{Robustness Assessment}

In order to ensure the safety of robot controllers, thorough evaluation is essential before real-world deployment. In particular, testing the controllers across a wide range of terrains or perturbations is required to investigate their weaknesses. Moreover, uncovering failure cases is crucial to further improve them. 

Recently, extensive research \cite{ding2023survey} has focused on identifying failure cases in Autonomous Vehicles (AVs), a field often referred to as \textit{Safety-Critical Driving Scenario Generation}. 
For instance, in KING \cite{hanselmann2022king}, a gradient-based safety-critical scenario generation procedure perturbs the trajectories of adversarial agents. The authors show that fine-tuning the policy on the generated data improves the controller's ability to avoid collisions. Similarly, in another study \cite{ding2021semantically}, explicit knowledge is used to guide the generation of safety-critical scenarios.

In the context of locomotion controllers, however, there is a lack of literature addressing this topic, and further research is needed to explore this field.
While some works propose methods to improve robustness, such as domain randomization \cite{tobin2017domain, li2023terrain}, research on assessing robustness itself remains limited and only a few studies have specifically focused on generating failure cases for legged robots.
The method presented in \cite{zhang2023generating} uses GANs to generate challenging and realistic terrains, aiming to create a benchmark for legged locomotion. However, it does not focus on identifying failure cases for specific controllers. In contrast, \cite{shi2024rethinking} and \cite{long2024learning} generate adversarial disturbances and apply them to the robot, and using these disturbances to improve the robustness of the controller.

\subsection{Quality-Diversity Algorithms}

QD algorithms are a class of optimization techniques that aim to generate a collection of diverse and high-quality solutions to a problem, rather than focusing solely on finding a single optimal solution. 

A well-known approach within QD algorithms is MAP-Elites \cite{mouret2015illuminating}, which arranges solutions in a grid-like archive based on descriptors that define their characteristics. Each cell in this grid stores the solution with the highest fitness (or objective) for a specific descriptor combination. In practice, MAP-Elites iteratively samples new solutions, evaluates them by their fitness and descriptors, and then places them in the corresponding cell of the archive if the cell is empty or contains a solution with lower fitness. These solutions, stored in the archive and called elites, are diverse and optimized within each cell, each representing a unique combination of descriptors.

QD algorithms often require substantial computational resources to explore a vast search space to generate a diverse set of high-quality solutions. Recent state-of-the-art algorithms, such as CMA-ME \cite{fontaine2020covariance} and CMA-MAE \cite{fontaine2023covariance} address this challenge by combining the MAP-Elites framework with the Covariance Matrix Adaptation Evolution Strategy (CMA-ES) \cite{hansen2001completely}. By leveraging CMA-ES's efficient covariance matrix adaptation, these methods explore the search space more effectively, enabling faster filling of the solution archive. Additionally, multiple emitters, which independently generate new solutions, are often employed to further accelerate the training process.

QD algorithms have been successfully applied in various fields, such as damage recovery for legged robots \cite{cully2015robots} and video games level generation \cite{fontaine2021illuminating}. 
Previous works effectively applied QD algorithms to explore failure cases in the video game Overcooked \cite{fontaine2021importance} and in human-robot collaboration with robotic arms \cite{fontaine2020quality}.
Additionally, CaDRE \cite{huang2024cadre} proposes using QD algorithms to generate safety-critical driving scenarios for AVs.

In the context of legged robot locomotion, the study in\cite{cully2015robots} uses a QD algorithm to learn a collection of walking gaits for a hexapod robot, allowing the robot to adapt to damage by quickly finding a compensatory behavior.
POET \cite{wang2019poet, wang2020enhanced} employs concepts similar to QD algorithms to grow a diverse population of environment-agent pairs in the Bipedal Walker environment of OpenAI Gym, in an open-ended manner.
In \cite{surana2023efficient}, the authors propose to learn a single walking policy by evolving a set of high-performing trajectory generator priors across a wide range of predefined terrains. Our method differs by focusing on assessing the robustness of a specific controller, rather than directly using a QD algorithm to learn one or several walking policies. It achieves this by evolving terrains to discover a wide range of failure cases.
%

\section{METHODOLOGY}
\label{methodology}

\begin{figure}[b]
    \centering
    \includegraphics[width=0.44\textwidth, trim={0 0 0 0.7cm},clip]{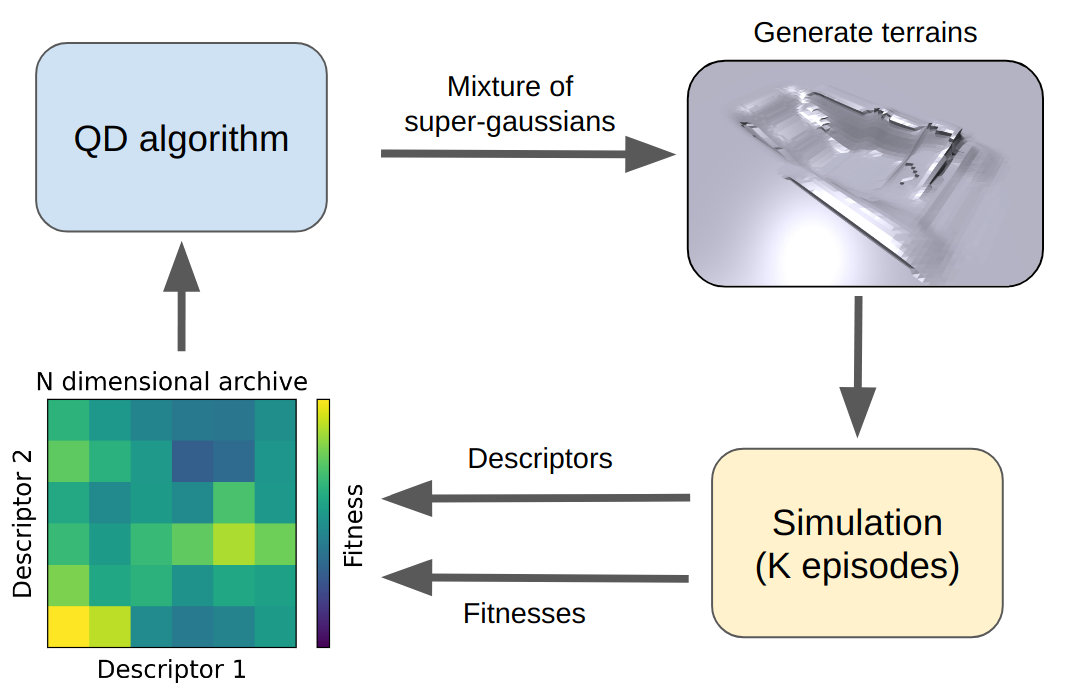}
    \caption{Illustration of the QD-based terrain generation framework.}
    \label{fig:framework}
\end{figure}

The proposed framework employs a QD algorithm to generate terrains designed to uncover diverse corner cases where a legged robot controller may fail.

This work utilizes a state-of-the-art QD algorithm, CMA-MAE, which fills an N-dimensional grid during training; each filled cell contains a solution, which in our case is a terrain. The goal is to optimize these terrains to be challenging for the target controller, with the descriptor space used to create the different cells representing a variety of failure modes. Fig. \ref{fig:framework} illustrates the proposed QD-based terrain generation framework. After training, the output of the framework is the archive, where optimized terrains can be selected based on their descriptors.

This section details the method as applied to the Cassie robot \cite{cassiegithub} and an RL-based controller, though it can be adapted for other types of legged robots and controllers.

The target controller used in this work is trained on hand-crafted terrains, such as stairs, sloped pyramids, random noise, and discrete obstacles, using the method and code provided in \cite{rudin2022learning} within Isaac Gym \cite{makoviychuk2021isaac}. The controller receives information about the robot's surroundings through measurements of the terrain around the robot's base. Each measurement corresponds to the distance between the terrain surface and the robot's base height, and this data is included in the observations. The full details of the controller are provided in \cite{rudin2022learning}, along with the accompanying code.
Two additional negative rewards were added to penalize high-foot contact forces and stumbles. Stumble is a boolean value that is true when one of the feet hits a vertical surface.

The focus of this work is on the task of having the robot traverse a rough terrain to reach a specific goal position on the other side of the terrain while consistently facing the goal. The robot must maintain a linear velocity of $v=0.75 m/s$. At each time step, the controller receives three input commands: linear velocity commands along the x and y axes, which are calculated based on $v$ and the robot's orientation relative to the goal, and an angular velocity command, which is calculated based on the angle error.

\subsection{Search space}

This subsection presents the search space used for the CMA-MAE algorithm, which the QD algorithm explores to generate diverse terrains. 
The search space must be kept relatively small to ensure efficient exploration and avoid excessive computational cost.

Each terrain has a total length of 16m (x-axis) and a width of 8m (y-axis), and the robot has a maximum of 20s to traverse it along the x-axis and reach the goal.
The terrain is represented as a 2D mixture of Super-Gaussians \cite{parent1992propagation}, which extend standard Gaussians by using a tunable exponent that allows the functions to have a flat top with sharper slopes. 
The mixture of $n$ Super-Gaussians is defined as follows:
\begin{align}
h(x, y) =& \sum_{i=1}^{n} w_i \cdot \exp\left(-\left(\frac{|x_{\text{rot},i}|}{\sigma_{x,i}}\right)^{2p_{x,i}}\right) \notag \\
&\quad \times \exp\left(-\left(\frac{|y_{\text{rot},i}|}{\sigma_{y,i}}\right)^{2p_{y,i}}\right),  \label{eq:supergaussian} \\
x_{\text{rot},i} =& \cos(\theta_i) \cdot (x - \mu_{x,i}) - \sin(\theta_i) \cdot (y - \mu_{y,i}), \\
y_{\text{rot},i} =& \sin(\theta_i) \cdot (x - \mu_{x,i}) + \cos(\theta_i) \cdot (y - \mu_{y,i}).
\end{align}

The coordinates $(x, y)$ are the point where the Super-Gaussian is evaluated, and $h$ corresponds to the height of the terrain at that point.
Each 2D Super-Gaussian is defined by eight parameters. The first two, $\mu_x$ and $\mu_y$, set the center, while $\sigma_x$ and $\sigma_y$ control the width along the $x$ and $y$ axes. $p_x$ and $p_y$ adjust the slope in the $x$ and $y$ directions. $\theta$ determines the rotation, and a scaling factor $w$ controls the overall size. 

In practice, the parameters are constrained to values between -1 and 1 to simplify the optimization process by ensuring a consistent range for all parameters. If they exceed these bounds, they are clipped to remain within the [-1, 1] range. Afterward, the Table \ref{table:gaussian_min_max} is used to rescale the parameters to their corresponding values before computing equation (\ref{eq:supergaussian}).

\begin{table}[t]
    \vspace{2mm}
    \centering
    \begin{tabular}{|c|c|c|}
    \hline
    \textbf{Parameter}    & \textbf{Min} & \textbf{Max} \\ \hline
    $\mu_x$(m)               & 6            & 10           \\ \hline
    $\mu_y$(m)               & 2            & 6            \\ \hline
    $\sigma_x$(m)            & 0.5          & 3            \\ \hline
    $\sigma_y$(m)            & 0.5          & 3            \\ \hline
    $p_x$                 & 1            & 4            \\ \hline
    $p_y$                 & 1            & 4            \\ \hline
    $\theta$(rad)              & $-\pi$       & $\pi$        \\ \hline
    $w$          & -0.25          & 0.25           \\ \hline
    \end{tabular}
    \caption{Minimum and maximum values for each super-gaussian parameter.}
    \label{table:gaussian_min_max}
    \vspace{-2mm}
\end{table}

\vspace{-15mm}

For each terrain, eight Super-Gaussians are used, meaning each terrain is defined by $8 \times 8 = 64$ parameters. Based on experimental results, this dimensionality has been found to be small enough for CMA-MAE to efficiently explore the search space and is sufficient to generate a large diversity of terrains.

\subsection{Descriptor space}

\begin{figure*}[t!]
    \centering
    \includegraphics[width=0.3\textwidth]{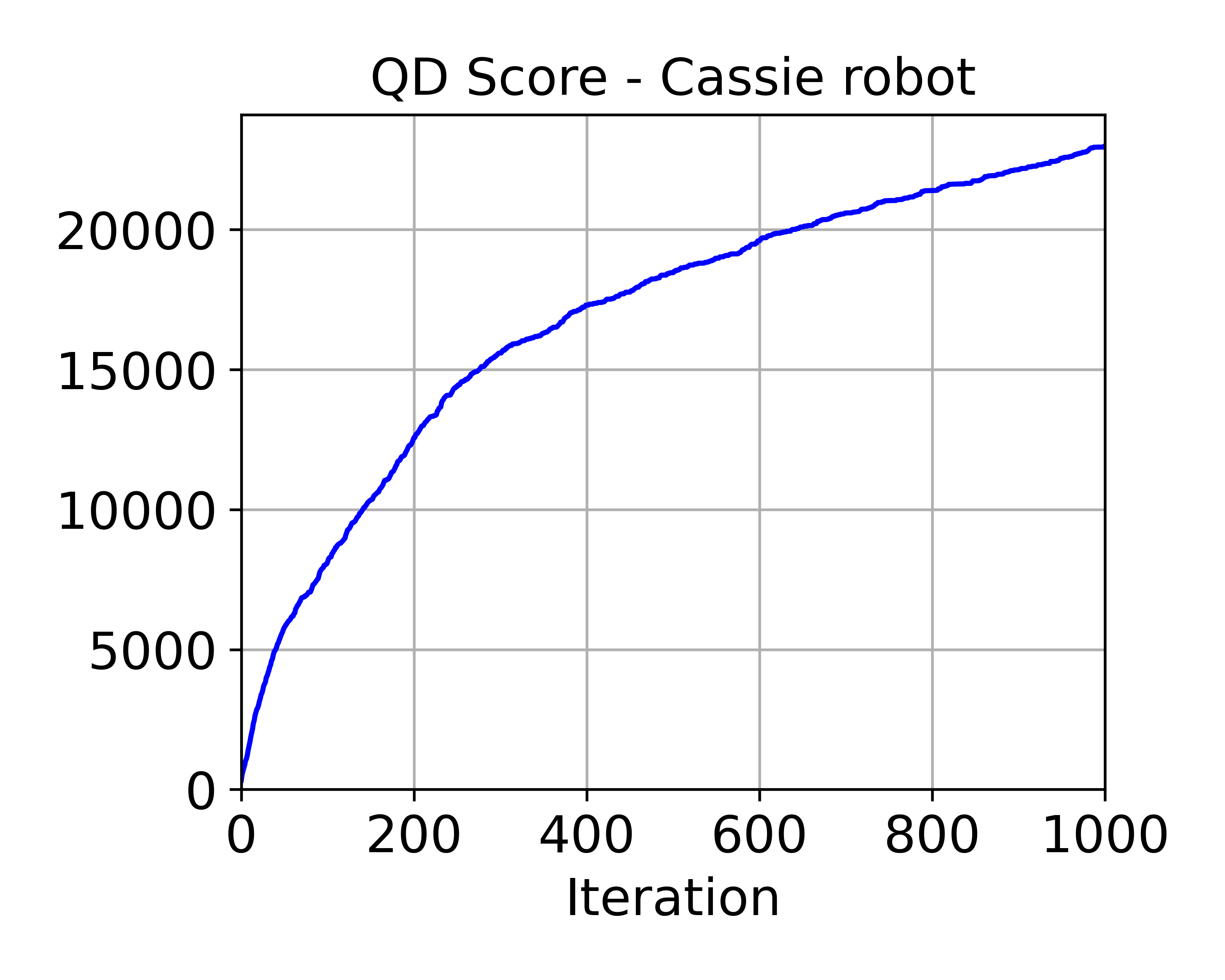}
    \hfill
    \includegraphics[width=0.3\textwidth]{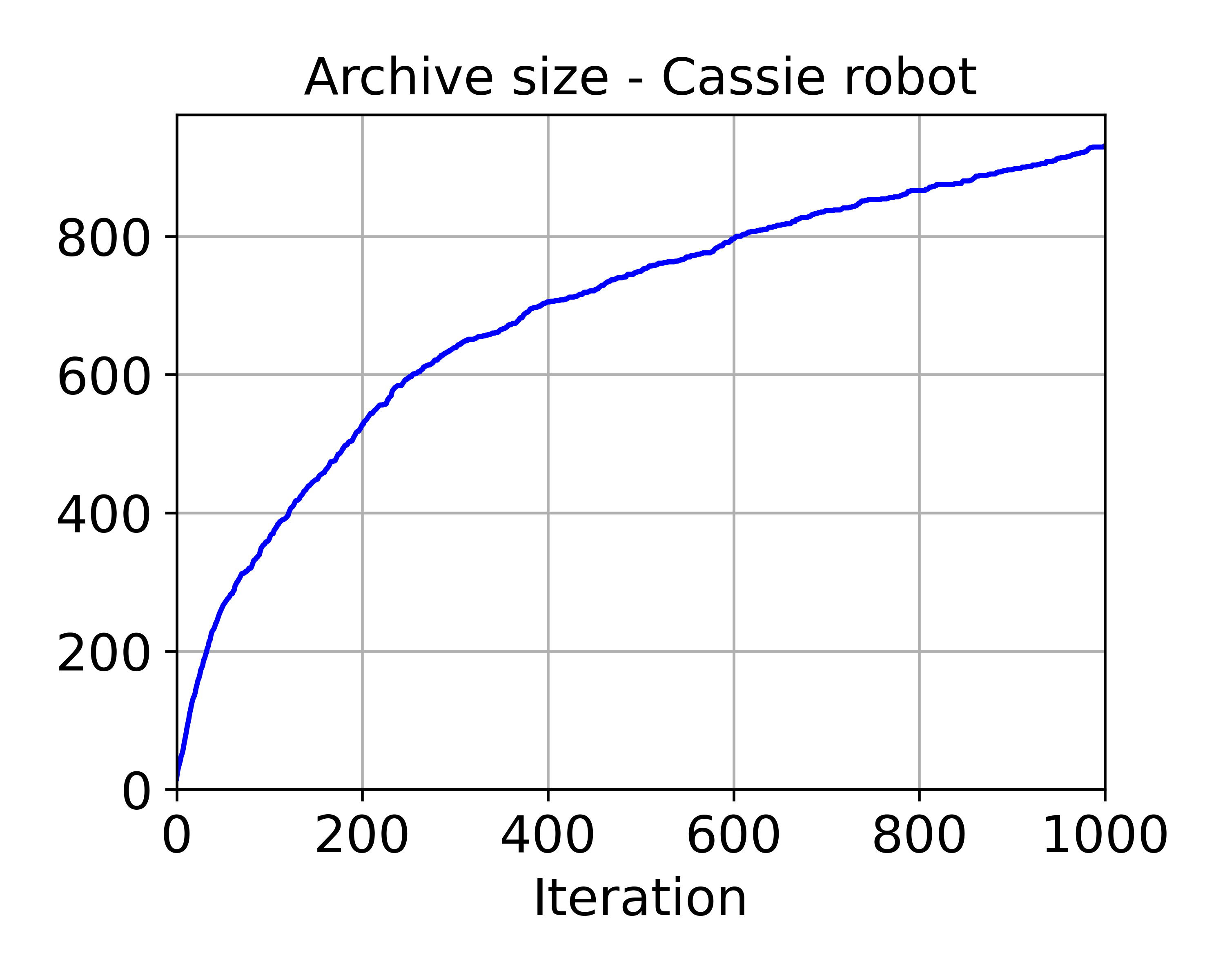}
    \hfill
    \includegraphics[width=0.3\textwidth]{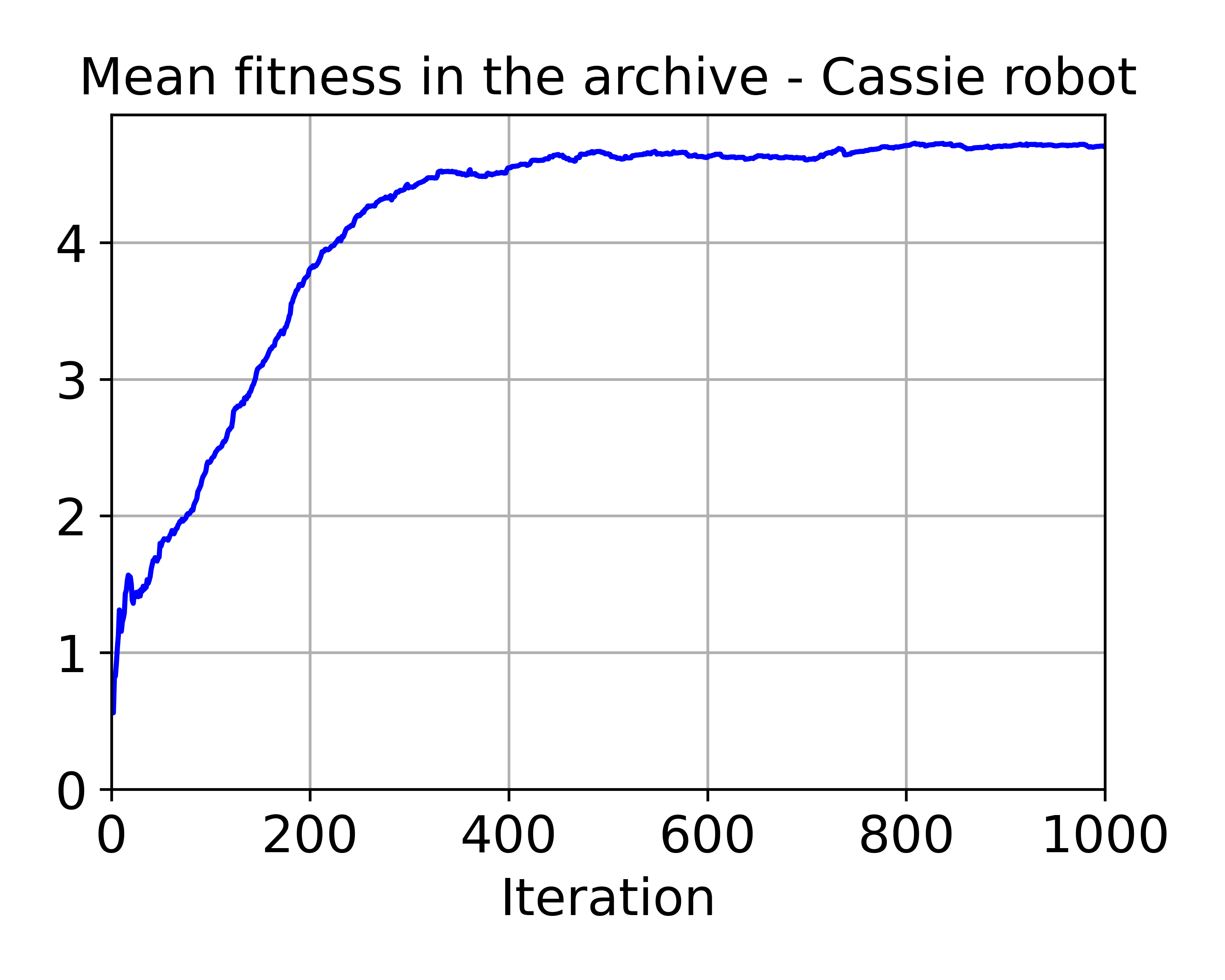}
    \caption{Evolution of QD score, archive size, and mean fitness for the Cassie robot.}
    \label{fig:cassie_training}
\end{figure*}

The descriptor space is used to define the N-dimensional grid that represents the diversity of solutions generated by CMA-MAE. In this work, the diversity of solutions is represented by the combination of penalties the controller receives when traversing the terrain, with $N=6$. 

Five penalties, referred to as \textit{raw penalties}, correspond to undesirable behaviors that legged robots should avoid. Three penalties are related to the robot's failure to follow the control inputs and are defined as errors in angular velocity and linear velocity in the $x$ and $y$ axes. The other two penalties account for high contact forces with the ground and stumbles, which must be minimized to prevent potential damage to the robot. During each episode, the \textit{raw penalties} are computed by summing the penalties at each time step, with a time step length of $dt=0.005s$. 
At the end of the episode, each raw penalty is normalized by dividing it by the sum of all five penalties, resulting in what we call \textit{ratio descriptors}. These ratio descriptors, defined specifically for our method, are used to define the descriptor space for the QD algorithm.
Each ratio descriptor has a value between 0 and 1, and the range is divided into 10 intervals (0.1 per range). 
A sixth descriptor is binary and indicates when there is a collision of the pelvis with the ground, which ends the episode. Overall, these descriptors ensure that each cell in the archive represents a distinct combination of failure modes.

Although the archive contains $10^5 \times 2=200000$ cells, most cells cannot be filled, as the five ratio descriptors must sum to 1. For this reason, the number of solutions in the archive is used to evaluate the results throughout the paper, rather than using the coverage metric, which measures the proportion of the archive that is filled, typically employed in QD algorithms. Additionally, it is necessary to scale each \textit{raw penalty} before calculating the ratio descriptors, ensuring that they are all of the same order of magnitude.

The proposed method is not limited to these penalties, and different penalties could be selected based on what needs to be avoided. 
Similarly, the number of intervals per descriptor can be adjusted: fewer intervals yield fewer terrains, while too many make the grid excessively large.

When evaluating the same controller multiple times on the same terrain, small changes in the initial state of the robot can cause significant variations in the robot's state and actions, in a similar way to the butterfly effect, ultimately leading to highly different descriptor values. This instability, which is a known issue in the QD field \cite{flageat2023uncertain}, is problematic because the generated terrains are less reliable for testing, with outcomes highly sensitive to minor variations or stochasticity. Thus, more reliable terrains should be generated, where similar descriptors are obtained even under slightly different conditions. To account for this problem, the robot is evaluated 20 times on each terrain with some randomness in the initial state, and the mean values of the raw descriptors are retained. For the binary collision descriptor, it is set to true if at least one collision occurs across the 20 episodes. A novel fitness function to address this issue, sharing some similarities with \cite{flageat2024exploring}, is provided in the next subsection.

\subsection{Fitness}

The fitness of each solution is defined as follows:
\begin{equation}
    f(S) = \sum_{j} \mathrm{mean}(\mathrm{pen}_j(S)) - \alpha \sum_{j} \mathrm{std}(\mathrm{pen}_j(S)) - \lambda \mathrm{u} (S),
\end{equation}
where $\mathrm{pen}_j(S)$ represents the raw penalty $j$ calculated over 20 episodes. The fitness function uses the mean of these penalties. It reflects the overall performance, i.e., how challenging a terrain is.
The standard deviation (STD) is subtracted to penalize inconsistencies across episodes and encourage the generation of terrains where the robot consistently fails in similar ways. $\mathrm{u}(S)$ is the non-collision rate, defined as the number of episodes without collision divided by the total number of episodes. This penalty is set to $0$ if no collision occurs at all. It further encourages the generation of terrains where the robot failures are consistent. $\alpha$ and $\lambda$ serve to adjust the penalties and are set to $1$ and $2$, respectively. Overall this formulation allows to maximize how challenging the terrains are while ensuring that the robot fails in a similar way when evaluated multiple times.

\section{EXPERIMENTS AND RESULTS}
\label{exp}

\subsection{Terrain Generation for the Cassie robot}

\subsubsection{Setup}

The pyribs library \cite{tjanaka2023pyribs} is used to implement the CMA-MAE algorithm. 
CMA-MAE runs using the proposed framework with 10 emitters, each having a population size of 20. Each solution is evaluated 20 times, resulting in 4,000 evaluations per iteration. By leveraging parallel evaluation on GPUs, as described in \cite{rudin2022learning}, each iteration takes approximately 30 seconds. With a total budget of 1,000 iterations, the training process takes about 8 hours using two 3070 Ti GPUs. 

\subsubsection{Results}

The QD score is a metric for QD algorithms introduced in \cite{pugh2015confronting}, which corresponds to the sum of the fitness of all solutions in the archive. Following previous works, and to avoid penalizing negative fitness values, an offset corresponding to the minimum possible fitness is added to each solution's fitness before computing the QD score. In this work, the offset is set to a value of $20$ as the minimum fitness encountered during training was close to $-20$. 

Fig. \ref{fig:cassie_training} shows the QD score, archive size, and mean fitness throughout the training. While the mean fitness of the solutions in the archive rapidly increases at the beginning and then stagnates around $4.7$, the QD score and archive size continue to slowly increase until the designated budget of 1000 iterations is reached. A total of 930 solutions were generated, 598 corresponding to solutions with a collision and 332 without collisions.

\begin{figure}[t]
    \centering
    \includegraphics[width=0.46\textwidth]{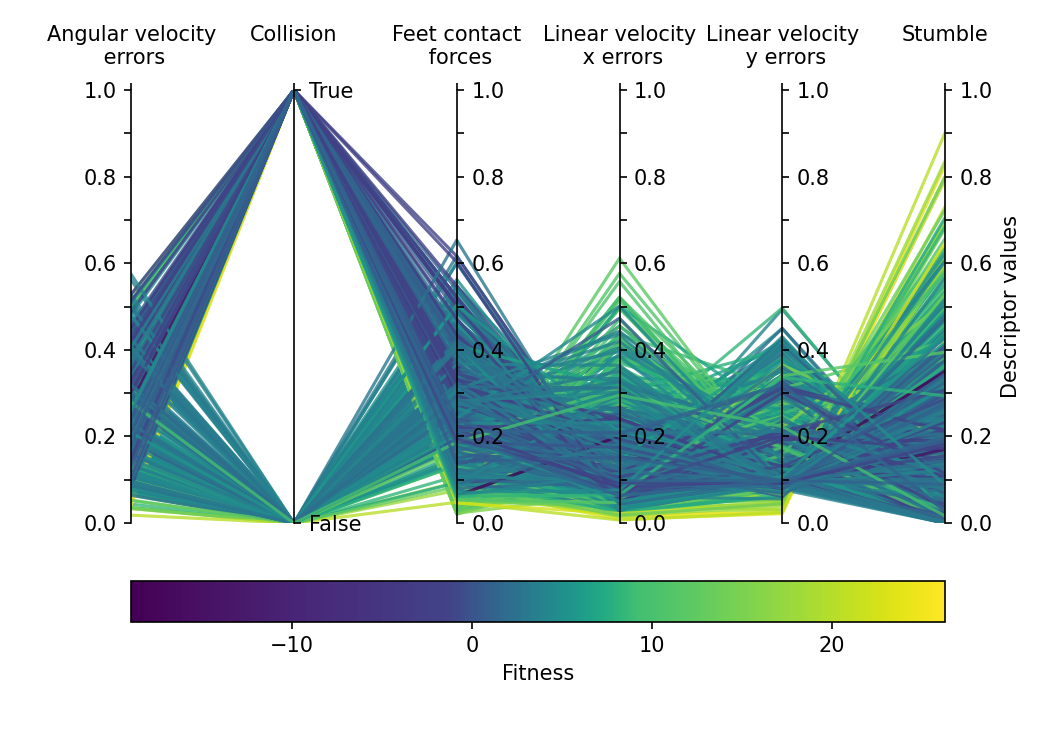}
    \caption{Trained archive of the Cassie robot. Each vertical axis represents one dimension of the archive. Each colored line represents a generated terrain in the archive and passes through the corresponding value for each descriptor.}
    \label{fig:parallel_archive}
\end{figure}

\begin{figure*}[t]
    \vspace{2mm}
    \centering
    
    \begin{subfigure}{0.24\textwidth}
        \centering
        \includegraphics[width=\textwidth]{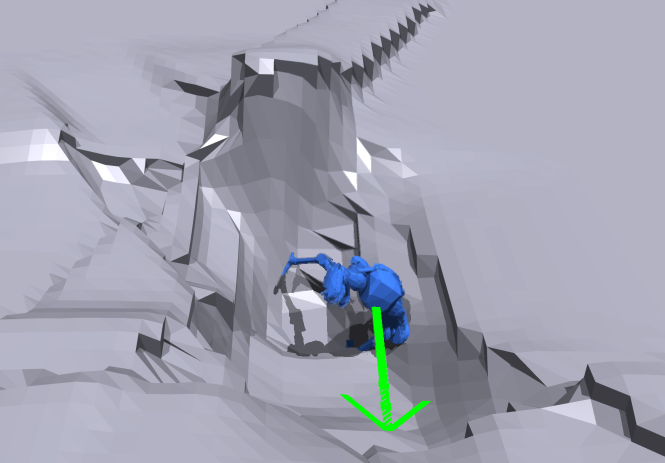}
        \caption{}
        \label{fig:cassie_ex_1}
    \end{subfigure}
    \hfill
    \begin{subfigure}{0.24\textwidth}
        \centering
        \includegraphics[width=\textwidth]{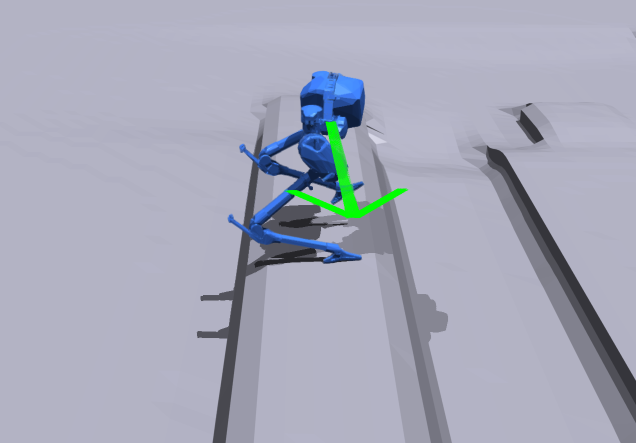}
        \caption{}
        \label{fig:cassie_ex_2}
    \end{subfigure}
    \hfill
    \begin{subfigure}{0.24\textwidth}
        \centering
        \includegraphics[width=\textwidth]{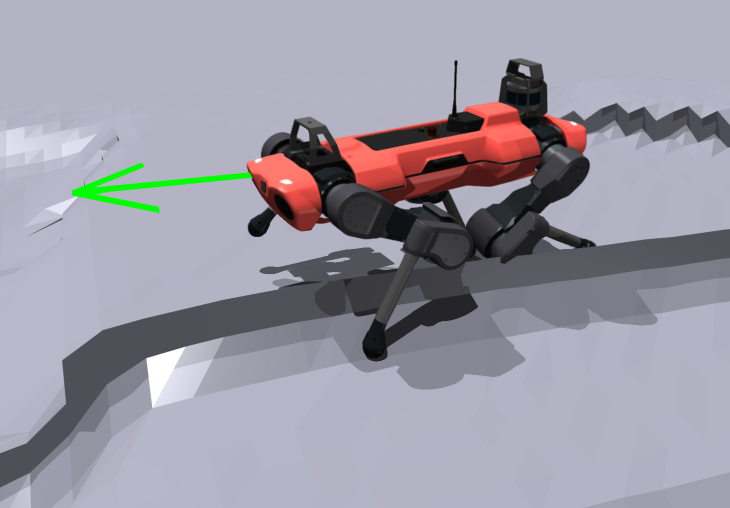}
        \caption{}
        \label{fig:anymal_ex_1}
    \end{subfigure}
    \hfill
    \begin{subfigure}{0.24\textwidth}
        \centering
        \includegraphics[width=\textwidth]{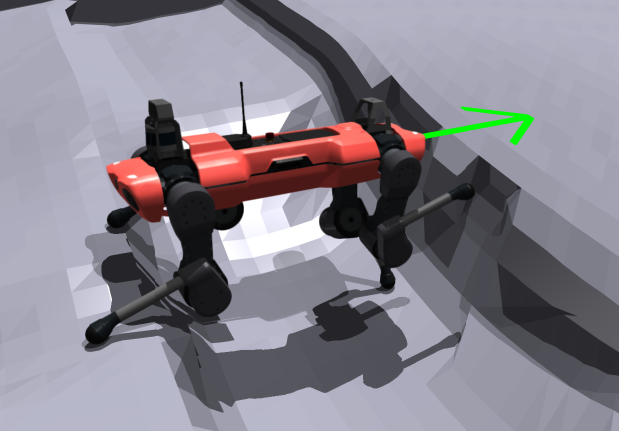}
        \caption{}
        \label{fig:anymal_ex_2}
    \end{subfigure}
    \caption{Examples of generated terrains. The green arrows represent the linear velocity commands in $x$ and $y$ directions, as well as the direction the pelvis should face.}
    \label{fig:env_examples}
    \vspace{-2mm}
\end{figure*}

The archive after training is presented in Fig. \ref{fig:parallel_archive}, allowing to visualize the solutions quickly. Cells with a ratio descriptor over $0.5$ are more rarely filled, as terrains where the controller mostly receives only one kind of penalty are uncommon. Many cells with ratio descriptor values below $0.5$ are filled, both with and without collisions.

The framework was able to produce some interesting and unexpected results. Fig. \ref{fig:env_examples}  shows some qualitative examples. Fig. \ref{fig:env_examples}(\subref{fig:cassie_ex_1}) depicts a generated terrain where the robot falls from a high platform and is unable to land properly, inducing a collision between the pelvis and the ground. The result shown in Fig. \ref{fig:env_examples}(\subref{fig:cassie_ex_2}) is particularly interesting. It corresponds to a terrain where the robot has a high heading error, resulting in a high value for the descriptor associated with angular velocity errors. This outcome revealed that the controller causes the robot to consistently head toward one side when traversing a narrow platform. This behavior was unexpected and, because it is only observed when the platform has a specific width, it may not have been discovered through testing on hand-crafted terrains alone.

\begin{figure}[ht]
    \centering
    \includegraphics[width=0.42\textwidth]{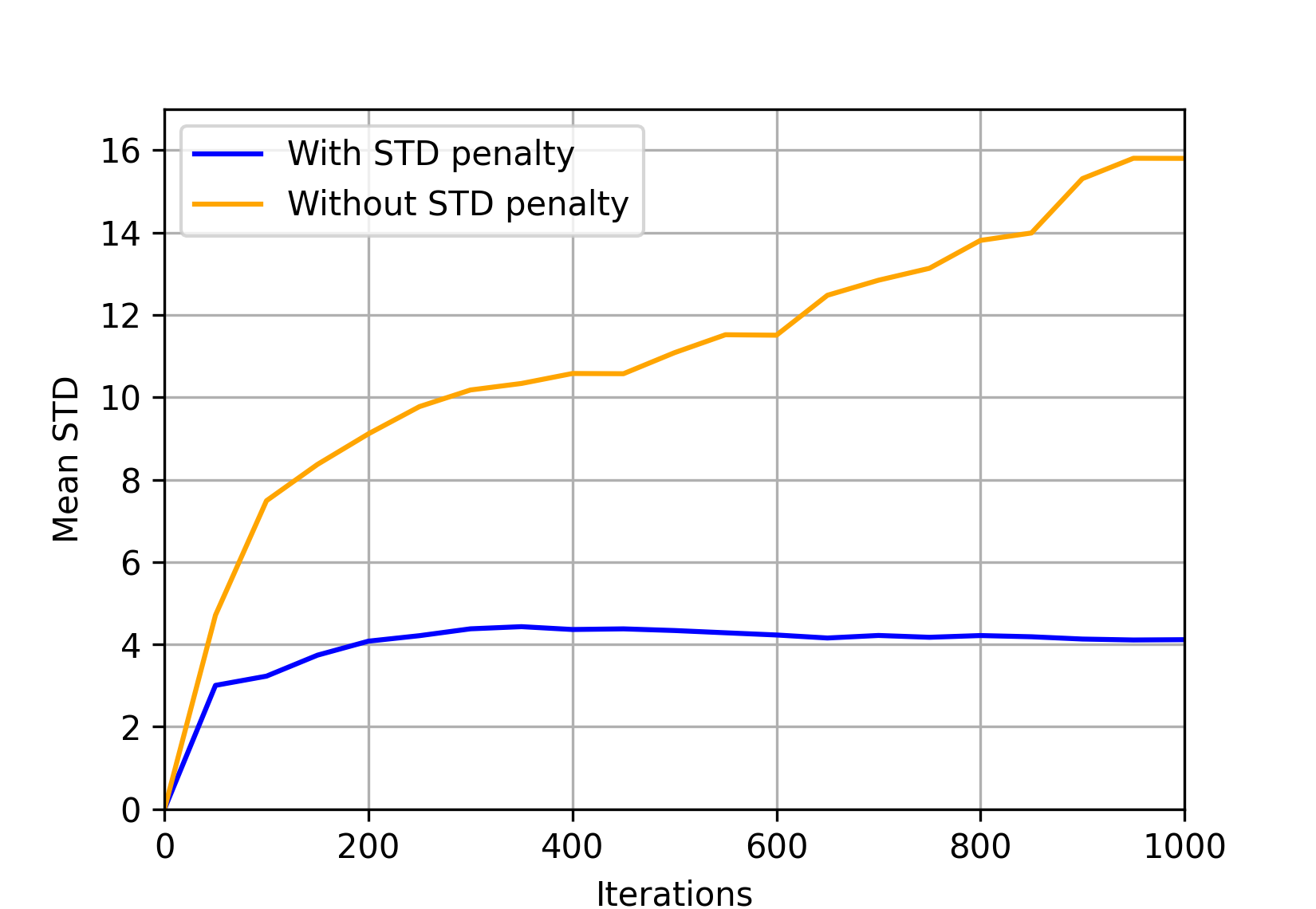}
    \caption{Mean STD computed on the terrains contained in each archive.}
    \label{fig:mean_std_plot}
\end{figure}

In order to evaluate the efficiency of the STD penalty, experiments are conducted by comparing a run using the proposed method with $\alpha=1$ with another run where $\alpha=0$, i.e., a baseline without the STD penalty. The archive is stored every 50 iterations, the STD is computed over 40 episodes for each terrain in the archive and the mean STD is reported in Fig. \ref{fig:mean_std_plot}. The mean STD when using the STD penalty is much lower than that of the baseline, indicating that the generated terrains are more reliable in the sense that the raw penalties received by the robot are more stable, thus demonstrating the effectiveness of the proposed method.

\subsection{ANYmal robot}

Our method is not specific to any particular robot type. Thus, additional experiments were conducted on a simulated ANYmal quadruped robot \cite{fankhauser2018anymal}.
Most of the experimental settings remain unchanged, with the only differences being how collisions are handled and the scaling of raw penalties.
For this experiment, collisions are considered both on the base, shanks, and thighs of the robot. Only collisions with the base are treated as catastrophic failures and terminate the simulation. In this setting, maximizing the number of collisions is appropriate since not all collisions terminate the simulation, therefore the number of collisions is used to compute a ratio descriptor. Thus, the dimensions of the archive correspond to six ratio descriptors, with no binary descriptor.

The framework successfully generated various challenging terrains, producing a total of 1,129 terrains in this experiment. For instance, in Fig. \ref{fig:env_examples}(\subref{fig:anymal_ex_1}), the robot walks along an obstacle, with its front left leg and rear right leg on opposite sides of the obstacle, leading to a large number of collisions with the rear left shank. In Fig. \ref{fig:env_examples}(\subref{fig:anymal_ex_2}), an obstacle blocks the robot, leaving it stuck in the configuration shown in the image.

\subsection{Controller Improvement}

\subsubsection{Setup}

An experiment is conducted to investigate whether the generated terrains can be used to improve the target controller in the case of an RL-based controller. This is achieved by starting with the policy used to generate the archive and fine-tuning it. \textit{C-G} is a Controller obtained by continuing the RL training on a dataset composed of half hand-crafted terrains and half generated (G) terrains taken randomly from the trained archive. A total of 750 terrains from the archive are utilized in this process.

Two other controllers are fine-tuned for each robot to serve as baselines. \textit{C-HC} is a Controller, trained solely on hand-crafted (HC) terrains and \textit{C-HR} is trained on a dataset consisting of $50\%$ of hand-crafted terrains and $50\%$ of hard random (HR) terrains. Random terrains are generated by randomly sampling the 64 Super-Gaussian parameters; however, this tends to produce terrains that are too easy to traverse. To address this problem, for hard-random terrains, the scaling factor of each Gaussian is randomly set to either its maximum or minimum value.
All 3 controllers for each robot are fine-tuned using the same method, with the same amount of episodes, the only difference being in the terrains they are trained on.

For the Cassie robot, 600 agents run simultaneously on a total of 100 terrains. The terrains are changed every 200 policy updates until reaching a total of 3000 policy updates. For the ANYmal robot, 1500 agents run on 500 terrains, which are changed every 400 policy updates, until reaching 1200 policy updates. Both settings result in the same number of simulation steps.
The chosen settings were the best settings for each robot, for \textit{C-G} and for two baselines.

\subsubsection{Results}

Each controller for each robot is tested on three different test sets, one containing only hand-crafted terrains, another with only hard-random terrains, and a third with terrains from the archive that were not used during training. Each dataset contains 180 terrains for the Cassie robot and 300 for the ANYmal robot, and the controllers are evaluated 20 times on each terrain, to account for the randomness.
Two metrics are used to evaluate the performance of the improved controllers. The first is the average reward and the second one is the success rate over all episodes of all terrains in a dataset. An episode is considered as a failure when the robot has a collision on its pelvis or body; otherwise, it is regarded as a success. This metric is used to observe the robot's most catastrophic failure mode.

\begin{table}[t]
\vspace{2mm}
\begin{minipage}{\linewidth}
\begin{center}
\caption{Average rewards with the Cassie robot ($\uparrow$)}
\begin{tabular}{|c|c|c|c|}
\hline
\diagbox{Test set}{Controller} & C-HC & C-HR & C-G (\textbf{Ours}) \\
\hline
Hand-crafted terrains & 32.7 & 29.7 & \textbf{32.9} \\
\hline
Hard-random terrains & 31.0 & 31.6 & \textbf{32.3} \\
\hline
Generated terrains & 25.7 & 27.6 & \textbf{29.9} \\
\hline
\end{tabular}
\label{table:rewards_cassie}
\end{center}

\begin{center}
\caption{Success rates (\%) with the Cassie robot ($\uparrow$)} 
\begin{tabular}{|c|c|c|c|}
\hline
\diagbox{Test set}{Controller} & C-HC & C-HR & C-G (\textbf{Ours}) \\
\hline
Hand-crafted terrains & 99.7 & 98.4 & \textbf{99.9} \\
\hline
Hard-random terrains & 92.8 & 98.2 & \textbf{99.0} \\
\hline
Generated terrains & 74.3 & 87.6 & \textbf{95.7} \\
\hline
\end{tabular}
\label{table:success_cassie}
\end{center}

\begin{center}
\caption{Average rewards with the ANYmal robot ($\uparrow$)}
\begin{tabular}{|c|c|c|c|}
\hline
\diagbox{Test set}{Controller} & C-HC & C-HR & C-G (\textbf{Ours}) \\
\hline
Hand-crafted terrains & \textbf{25.6} & 24.8 & 25.4 \\
\hline
Hard-random terrains & 22.7 & 24.9 & \textbf{25.6} \\
\hline
Generated terrains & 16.6 & 20.1 & \textbf{23.3} \\
\hline
\end{tabular}
\label{table:rewards_anymal}
\end{center}

\begin{center}
\caption{Success rates with the ANYmal robot ($\uparrow$)}
\begin{tabular}{|c|c|c|c|}
\hline
\diagbox{Test set}{Controller} & C-HC & C-HR & C-G (\textbf{Ours}) \\
\hline
Hand-crafted terrains & 97.1 & 97.6 & \textbf{98.0} \\
\hline
Hard-random terrains & 86.5 & 95.4 & \textbf{97.6} \\
\hline
Generated terrains & 59.8 & 87.6 & \textbf{91.4} \\
\hline
\end{tabular}
\label{table:success_anymal}
\end{center}
\end{minipage}
\vspace{-2mm}
\end{table}

Tables \ref{table:rewards_cassie} and \ref{table:success_cassie} present the results for the Cassie robot. Our method, \textit{C-G}, outperforms the baselines on both hard-random terrains and generated terrains. 
Even though the generated terrains used for training and for testing are not the same, they may present similarities, especially if their location in the descriptor space is close, creating a risk of overfitting that may not be detected during the tests. However, the performances of \textit{C-G} on hard-random terrains show that the finetuned policy is able to generalize to other terrains as well.
\textit{C-G} also achieves comparable performances to \textit{C-HC} on hand-crafted terrains.
Similar results can be observed with the ANYmal robot in Tables \ref{table:rewards_anymal} and \ref{table:success_anymal}.

\begin{figure}[t]
    \vspace{2mm}
    \centering
    \begin{subfigure}{0.48\columnwidth} 
        \centering
        \includegraphics[width=\textwidth]{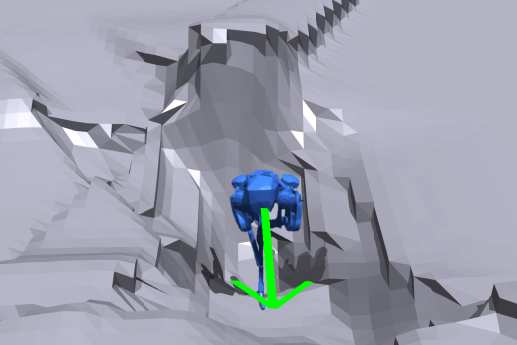}
        \caption{}
        \label{fig:cassie_ex_2_after}
    \end{subfigure}
    \hspace{0em} 
    \begin{subfigure}{0.48\columnwidth} 
        \centering
        \includegraphics[width=\textwidth]{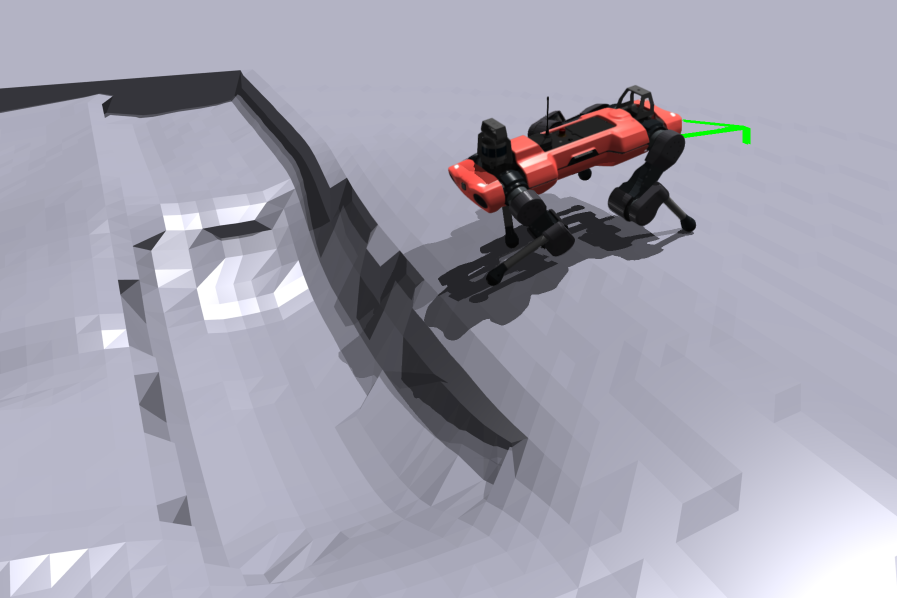}
        \caption{}
        \label{fig:anymal_ex_2_after}
    \end{subfigure}
    
    \caption{Examples showing improvements of the controllers using our method.}
    \label{fig:retraining_examples}
    \vspace{-2mm}
\end{figure}

Fig. \ref{fig:retraining_examples} displays some qualitative results. While the target controller was unable to land properly on the terrain shown in Fig. \ref{fig:env_examples}(\subref{fig:cassie_ex_1}), \textit{C-G} goes around the highest platform and falls from a lower height, avoiding a catastrophic failure, as shown in Fig. \ref{fig:retraining_examples}(\subref{fig:cassie_ex_2_after}). In Fig. \ref{fig:retraining_examples}(\subref{fig:anymal_ex_2_after}), the \textit{C-G} of the ANYmal robot is able to pass over the obstacle where the target controller was previously stuck (Fig. \ref{fig:env_examples}(\subref{fig:anymal_ex_2})).

These results demonstrate the effectiveness of the proposed method. By focusing on generating terrains where the controller receives diverse combinations of penalties, the controller is provided with relevant feedback that enables it to improve its performance.

\section{CONCLUSION}
\label{conclusion}

This work presents a QD-based framework that generates terrains to discover various weaknesses of legged robot controllers.
The proposed framework is applied to simulations of both bipedal and quadruped robots, successfully uncovering various weaknesses in the controllers, as demonstrated by the results of the experiments. By leveraging multi-parallelism \cite{rudin2022learning}, the framework is able to efficiently explore a wide range of terrains, generating satisfying results in a reasonable amount of time despite being computationally expensive.
Additional experiments demonstrate that the generated terrains can be utilized to improve the controller.
We hope this paper paves the way for the development of methods to identify corner cases in legged robots, ultimately leading to more robust controllers.

While promising, the proposed approach does present a few limitations. Notably, the current framework does not guarantee that the generated terrains are traversable by the robot. Additionally, the generated terrains are not always realistic, a limitation we plan to address in future work. Indeed, thanks to the use of multi-parallelism, which enables high computational capacity, the framework could be scaled up to handle more complex and realistic terrains, including varying friction and stiffness, helping to uncover all potential corner cases.
Furthermore, advancements in QD algorithms could further benefit the framework, particularly by incorporating a surrogate function, as recently suggested in the QD literature, to replace direct terrain evaluations and significantly enhance the method's efficiency.

\bibliographystyle{IEEEtran}
\bibliography{manual}

\end{document}